\documentclass[10pt,twocolumn,letterpaper]{article}

\usepackage{cvpr}              % To produce the CAMERA-READY version
\usepackage{multirow}
\usepackage{graphicx}
\usepackage{amsmath}
\usepackage{amssymb}
\usepackage{booktabs}
\usepackage{bm}

\usepackage{pifont}% http://ctan.org/pkg/pifont
\usepackage{xcolor}         % colors
\usepackage{colortbl}
\usepackage{makecell}
\usepackage{multirow}
\definecolor{demphcolor}{RGB}{144,144,144}
\definecolor{mygray}{gray}{0.95}
\usepackage{subcaption}

\newcommand{\cmark}{\ding{51}}%
\newcommand{\xmark}{\ding{55}}%

\definecolor{cvprblue}{rgb}{0.21,0.49,0.74}
\usepackage[pagebackref,breaklinks,colorlinks,allcolors=cvprblue]{hyperref}

\def\paperID{*****} % *** Enter the Paper ID here
\def\confName{CVPR}
\def\confYear{2025}

\title{Factorized Visual Tokenization and Generation}

\author{
Zechen Bai\textsuperscript{1},
Jianxiong Gao\textsuperscript{2},
Ziteng Gao\textsuperscript{1},
Pichao Wang\textsuperscript{3},
Zheng Zhang\textsuperscript{3},
Tong He\textsuperscript{3},
Mike Zheng Shou\textsuperscript{1}\thanks{Corresponding author}
\\
\textsuperscript{1}{Show Lab, National University of Singapore} \quad
\textsuperscript{2}{Fudan University} \quad
\textsuperscript{3}{Amazon}
}

\begin{document}
\maketitle

\begin{abstract}
Visual tokenizers are fundamental to image generation.
They convert visual data into discrete tokens, enabling transformer-based models to excel at image generation.
Despite their success, VQ-based tokenizers like VQGAN face significant limitations due to constrained vocabulary sizes.
Simply expanding the codebook often leads to training instability and diminishing performance gains, making scalability a critical challenge.
In this work, we introduce Factorized Quantization (FQ), a novel approach that revitalizes VQ-based tokenizers by decomposing a large codebook into multiple independent sub-codebooks.
This factorization reduces the lookup complexity of large codebooks, enabling more efficient and scalable visual tokenization.
To ensure each sub-codebook captures distinct and complementary information, we propose a disentanglement regularization that explicitly reduces redundancy, promoting diversity across the sub-codebooks.
Furthermore, we integrate representation learning into the training process, leveraging pretrained vision models like CLIP and DINO to infuse semantic richness into the learned representations.
This design ensures our tokenizer captures diverse semantic levels,
leading to more expressive and disentangled representations.
Experiments show that the proposed FQGAN model substantially improves the reconstruction quality of visual tokenizers, achieving state-of-the-art performance.
We further demonstrate that this tokenizer can be effectively adapted into auto-regressive image generation.
\url{https://showlab.github.io/FQGAN}
\end{abstract}

\section{Introduction}

\begin{figure}[t!]
    \centering
    \includegraphics[width=0.9\linewidth]{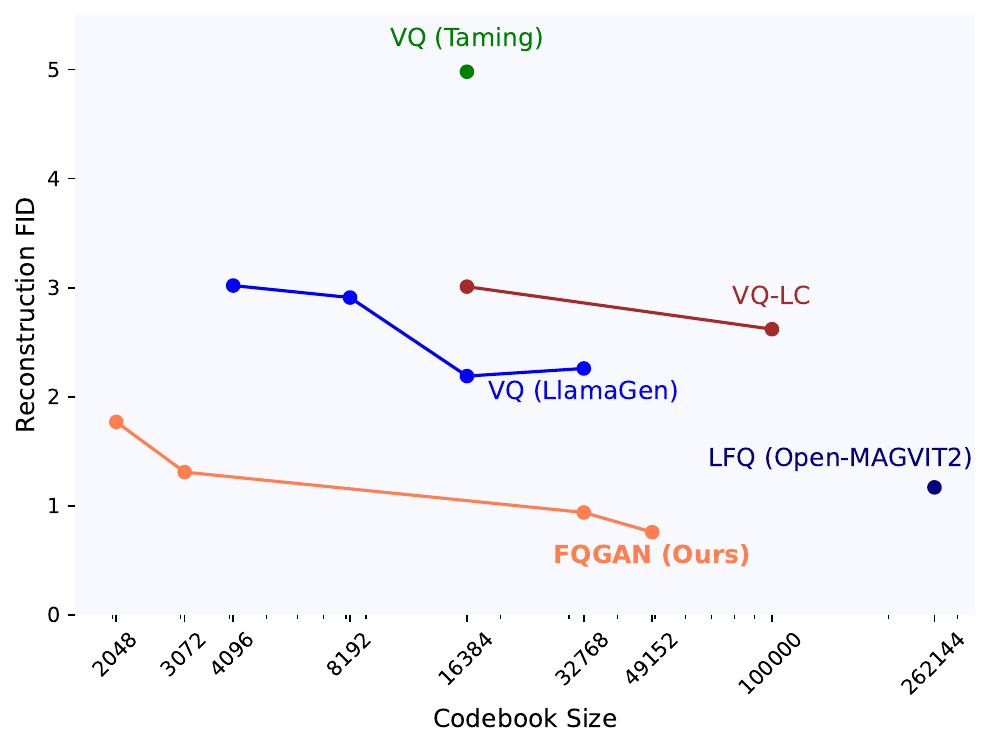}
    \vspace{-5pt}
    \caption{
    Performance comparison of popular tokenizers at various codebook sizes, including VQ (Taming)~\cite{vqgan}, VQ (LlamaGen)~\cite{llamagen}, VQ-LC~\cite{zhu2024scaling_vq}, LFQ (OpenMAGVIT2)~\cite{openmatvit2}, and FQGAN.
    Lower rFID values indicate better performance.
    }
    \vspace{-10pt}
    \label{fig:teaser}
\end{figure}

In recent years, the success of discrete token-based approaches in natural language processing~\cite{brown2020language, radford2021learning} has sparked growing interest in discrete image tokenization and generation~\cite{vqgan,vit-vqgan,rqvae}.
Visual tokenizers play a crucial role by converting image data into discrete tokens, thereby enabling the application of powerful transformer-based generative models.
The quality of visual tokenization significantly impacts high-fidelity image reconstruction and generation.

Popular visual tokenizers, such as VQGAN~\cite{vqgan}, adopt an encoder-quantizer-decoder structure, where the quantizer converts the latent feature into discrete tokens via vector quantization (VQ).
These approaches have shown remarkable performance on image reconstruction and generation~\cite{rqvae,llamagen,maskgit}.
Despite these successes, visual tokenization involves inherently lossy compression, especially compared to continuous encoding, since visual data is naturally continuous and complex.
A common strategy to address this is to enlarge the codebook, enhancing its capacity to approximate continuous representations.
However, traditional VQ-based models are constrained by codebook size.
Existing research~\cite{llamagen,zhu2024scaling_vq} indicates that increasing codebook sizes beyond 16,384 can lead to training instability, such as low codebook utilization and performance saturation.

Recent works have proposed innovative strategies to address these limitations.
For example, FSQ~\cite{fsq} and LFQ~\cite{magvit2} are introduced to eliminate the need for an explicit codebook, achieving state-of-the-art reconstruction quality using a massive codebook size.
Among VQ tokenizers, VQGAN-LC~\cite{zhu2024scaling_vq} employs pre-trained feature clusters to help stabilize training with larger codebooks.
Nevertheless, VQ tokenizers still exhibit inferior performance to LFQ ones and, more importantly, the inherent challenges of VQ remain unresolved.
Large codebooks complicate quantization by necessitating the calculation of pairwise distances between encoder outputs and all codebook entries, followed by an \texttt{argmin()} operation to select the nearest code.
As the codebook size increases, the lookup process becomes more computationally expensive and unstable, leading to inconsistent results.

To tackle these challenges, we draw inspiration from the divide-and-conquer principle, breaking down a complex problem into smaller, more manageable components to enhance both stability and performance.
We propose a novel factorized codebook design, wherein a large codebook is split into several smaller, independent sub-codebooks.
This factorization simplifies the tokenization process, improving the stability of the quantization.
By combining entries from multiple sub-codebooks, we construct a more expressive and scalable representation space.
It provides greater flexibility for capturing image features at varying levels of granularity, improving overall tokenization quality.

However, to fully harness this expressiveness and ensure that each sub-codebook contributes uniquely to the representation, it is essential to disentangle the learned features.
Factorizing the codebook alone is insufficient unless each sub-codebook learns to capture unique and complementary information.
To address this, we introduce a disentanglement regularization mechanism that enforces orthogonality between sub-codebooks, encouraging each sub-codebook to focus on distinct aspects of the visual data, such as spatial structures, textures, colors, etc.
This is akin to having different specialists analyzing various aspects of an image, ultimately resulting in a richer and more comprehensive representation.

To further enhance the specialization of the sub-codebooks, we integrate representation learning as an essential part of the training framework.
By seamlessly weaving into the training objective, the sub-codebook is guided to capture semantically meaningful features that contribute to the overall representation.
Traditional reconstruction objectives often lead to overfitting on high-variance visual details, which results in features that lack semantic meaning for perception tasks~\cite{lecun_recon}.
Our representation learning objective addresses this issue by guiding the factorized codebooks to learn robust, semantically rich features capable of generalizing beyond simple reconstruction.
Specifically, by leveraging different vision backbones, e.g., CLIP~\cite{radford2021learning} and DINOv2~\cite{dinov2}, the sub-codebooks essentially learn to establishes a complementary hierarchy of semantics: low-level structures (e.g., edges), mid-level details (e.g., textures), and high-level concepts (e.g., abstract appearance).

By seamlessly integrating factorized codebook design, disentanglement regularization, and representation learning objectives, our visual tokenizer captures a diverse and rich set of features.
This holistic approach greatly enhances reconstruction quality, as each sub-codebook learns to represent different aspects of the image in a balanced and semantically meaningful way.
Leveraging these three core innovations, our visual tokenizer achieves state-of-the-art performance in discrete image reconstruction, as illustrated in Fig.~\ref{fig:teaser}.
Additionally, we extend our analysis to auto-regressive (AR) generation tasks.
Unlike conventional tokenizers that produce a single token per image patch, our tokenizer encodes each patch into multiple tokens, resulting in a richer and more expressive representation.
Drawing inspiration from related works on handling extremely large codebooks~\cite{openmatvit2} and multi-code~\cite{rqvae}, we design a factorized AR head that predicts sub-tokens for each patch, adapting our tokenizer effectively for downstream image generation.

In summary, our contributions include:
\begin{itemize}
\item A novel factorized quantization design that revitalizes VQ-based tokenizers, achieving state-of-the-art performance on discrete image reconstruction.
\item Introduction of disentanglement and representation learning mechanisms that enable diverse and semantically meaningful codebook learning.
\item Demonstration that our tokenizer enhances downstream AR models, improving image generation quality on the ImageNet benchmark.
\end{itemize}

\section{Related Work}

\subsection{Visual Tokenizers}

Visual tokenizers map images into a latent space for downstream tasks, such as visual understanding~\cite{llava} and visual generation~\cite{vqgan}.
Visual tokenizers generally fall into two categories: continuous feature-based models and discrete token-based models.
We mainly discuss the discrete ones as they are closely related to our work.
Popular visual tokenizers, exemplified by VQGAN~\cite{vqgan}, use an encoder-quantizer-decoder structure: the encoder maps image data to a latent space, the quantizer transforms this representation into discrete tokens using vector quantization, and the decoder reconstructs the image from these tokens.
Building on VQGAN framework, numerous works~\cite{rqvae,vit-vqgan,zheng2022movq,hq_vae,fsq} have been developed to improve performance from various perspectives.
ViT-VQGAN~\cite{vit-vqgan} upgrades the encoder-decoder architecture from a CNN-based network to a transformer-based one.
RQ-VAE~\cite{rqvae} proposes modeling residual information using multiple codes to capture finer details.

Despite advancements, VQ tokenizers still struggle with the critical challenge of limited codebook size.
Research~\cite{llamagen,zhu2024scaling_vq} indicates that expanding the codebook size beyond 16,384 can lead to performance saturation or even degradation.
This issue is often accompanied by a low usage rate of the large codebook.
To address this, FSQ (Finite Scalar Quantization)~\cite{fsq} and LFQ (Lookup Free Quantization)~\cite{magvit2} are proposed to eliminate the need for an explicit codebook and significantly alleviates this issue.
Within the VQ family, VQGAN-LC~\cite{zhu2024scaling_vq} uses pre-trained feature clusters to implicitly regularize the large codebook, helping to maintain a higher usage rate.
This work suggests that semantic information can benefit visual tokenizers, a concept further explored in recent studies~\cite{gu2024rethinking,lg_vq,vila_u,image_understand_tok,v2l_token,yu2024spae}.
For instance, VILA-U~\cite{vila_u} demonstrates that a pre-trained vision model can be fine-tuned into a visual tokenizer while preserving its semantic capabilities.
LG-VQ~\cite{lg_vq} and VQ-KD~\cite{image_understand_tok} show that incorporating language supervision or image understanding models can improve visual tokenizers.
A concurrent work, ImageFolder~\cite{imagefolder}, proposes a folded quantization approach, improving the image reconstruction performance by a large margin.
Our work, as part of the VQ family, aims to revitalize VQ tokenizers by addressing the large codebook problem through factorized quantization and leveraging semantic supervision.
A more detailed discussion with related works is provided in the Appendix.

\subsection{Auto-regressive Visual Generation}

Auto-regressive visual generation uses a next-token prediction approach to sequentially generate images or videos.
VQGAN~\cite{vqgan}, a pioneering model, utilizes a transformer to predict tokens sequentially.
RQ-VAE~\cite{rqvae} extends VQGAN by incorporating a residual tokenization mechanism and adding an AR transformer head to predict residual tokens at a finer depth.
LlamaGen~\cite{llamagen} extends the VQGAN transformer architecture to the Llama~\cite{llama2} framework, demonstrating promising scaling behaviors.
VAR~\cite{var} extends next-token prediction to next-scale prediction, reducing auto-regressive steps and enhancing performance.
Open-MAGVIT2~\cite{openmatvit2}, similar to LlamaGen~\cite{llamagen}, adopts a Llama-style auto-regressive transformer as its backbone.
To manage an extremely large codebook, it predicts two sub-tokens during the AR generation phase and composes them to obtain the original code.
It also employs an RQ-like architecture, termed intra-block, to predict sub-tokens.
In this work, our factorized codes share similarities with RQ-VAE~\cite{rqvae} and Open-MAGVIT2~\cite{openmatvit2}, specifically in predicting multiple tokens at each AR step.
Consequently, we use a factorized AR head atop the AR backbone to predict sub-tokens for each patch.

\section{Method}

\begin{figure*}[t!]
    \centering
    \includegraphics[width=0.97\linewidth]{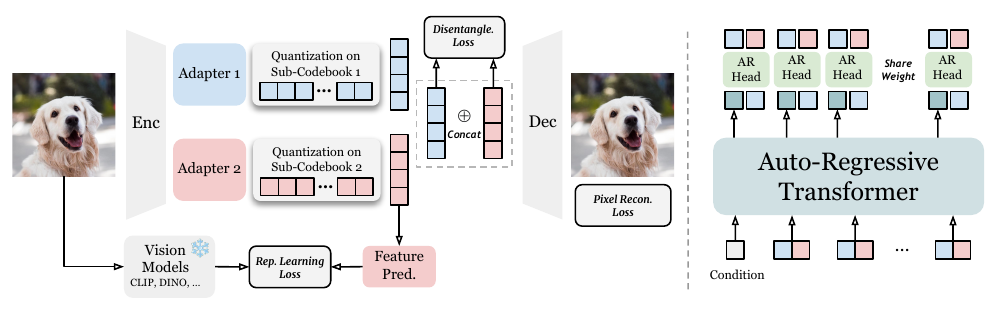}
    \caption{
    Illustration of the our method.
    The left part shows FQGAN-Dual, the factorized tokenizer design in an example scenario when $k=2$.
    This framework is extendable to factorization of more codebooks.
    The right part demonstrate how we leverage an additional AR head to accommodate the factorized sub-codes based on standard AR generative transformer.
    }
    \label{fig:method}
\end{figure*}

\subsection{Preliminary}
VQGAN~\cite{vqgan} employs a learnable discrete codebook $\mathbf{C}\in \mathbb{R}^{K\times D}$ to represent images, where $K$ is the codebook size while $D$ is the dimensionality of the codes.
Given an input image $x$, the encoder transforms it into a latent feature $h=\text{Enc}(x)$.
Then, the closest codebook entry for each patch is retrieved from the codebook to serve as the quantized representation:
\begin{equation}
q^{i} = \text{Quant}(h^{i}, \mathbf{C}) := \underset{c^j \in \mathbf{C}}{\arg \min }\left\|h^{i}-c^j\right\|,
\end{equation}
where $h^{i} \in \mathbb{R}^D$, $c^{j} \in \mathbb{R}^D$, $q^{i} \in \mathbb{R}^D$ denotes the encoded latent feature at patch $i$, codebook entry, and quantized feature at patch $i$, respectively.
After that, VQGAN uses a decoder to reconstruct the image $\hat{x}=\text{Dec}(q)$.
The training objective of VQGAN is to identify the optimal compression model of $\{\text{Enc}, \text{Dec}, \mathbf{C}\}$, involving the following loss:
\begin{equation}
    \mathcal{L}_{\text{VQGAN}} = \mathcal{L}_{\text{rec}} + \mathcal{L}_{\text{VQ}}+\mathcal{L}_{\text{perceptual}}+\mathcal{L}_{\text{GAN}},
\end{equation}
where $\mathcal{L}_{\text{rec}}$ denotes the pixel reconstruction loss between $x$ and $\hat{x}$.
$\mathcal{L}_{\text{VQ}}$ denotes the codebook loss that pulls the latent features $h$ and their closest codebook entries $q$ closer.
$\mathcal{L}_{\text{perceptual}}$ denotes the perceptual loss between $x$ and $\hat{x}$ by leveraging a pre-trained vision model~\cite{lpips}.
$\mathcal{L}_{\text{GAN}}$ introduces an adversarial training procedure with a patch-based discriminator~\cite{patchgan} to calculate the GAN loss.
As these losses are widely adapted in most VQ tokenizer designs~\cite{vqgan,rqvae,llamagen,zhu2024scaling_vq}, we omit the detailed definitions for simplicity.

\subsection{Factorized Quantization}
Despite the remarkable performance achieved by the classical VQGAN model, it is known to suffer from unstable training and low codebook usage rate when increasing the codebook size.
One prominent issue is the unstable lookup process among a large set of embeddings.
To alleviate this, we propose a factorized quantization approach that decomposes a singe large codebook $\mathbf{C}$ into $k$ small sub-codebooks.
The main framework is illustrated in Fig.~\ref{fig:method}.

\noindent\textbf{Encoder.}
We regard the original VQGAN encoder as a base feature extractor.
On top of that, $k$ feature adapters are introduced to transform the base image features into their respective feature space.
Formally,
\begin{equation}
    h_{base} = \text{Enc}(x),
\end{equation}
\begin{equation}
    h_{1}, h_{2}, ..., h_{k} = F_{1}(h_{base}), F_{2}(h_{base}), ..., F_{k}(h_{base}),
\end{equation}
where $F_{1}, ..., F_{k}$ are the adapters for each factorized branch.

\noindent\textbf{Quantizer.}
Our method maintain a unique codebook for each factorized branch.
After extracting the branch-specific features, the quantization process is conducted at each codebook independently.
Formally,
\begin{equation}
\label{eq:quant}
    q_{1}, ..., q_{k} = \text{Quant}(h_{1}, \mathbf{C_{1}}), ..., \text{Quant}(h_{k}, \mathbf{C_{k}}),
\end{equation}
where $\mathbf{C_{1}}, ..., \mathbf{C_{k}}$ are the factorized sub-codebooks.

\noindent\textbf{Decoder.}
Given the quantized feature from each sub-codebook, we employ a simple yet effective aggregation approach that concatenates them along the latent (channel) dimension.
After that, the aggregated features are fed into the pixel decoder, which is inherited from the VQGAN model.
Formally,
\begin{equation}
    \hat{x} = \text{Dec}(\left[q_{1}; q_{2}; ...; q_{k}\right]),
\end{equation}
where ``$;$" denotes the concatenation operation.

The factorized quantization design presents several appealing properties.
First, the factorized and parallelized lookup process greatly alleviates the lookup instability in a single large codebook.
Second, maintaining factorized sub-codebooks and independent feature adapters allow the model to learn more diverse features.
Lastly, the code aggregation before decoding essentially builds a super large \textit{conceptual} codebook with a size of $|\mathbf{C_i}|^k$.
E.g., suppose $k=2$, $|\mathbf{C_1}|=|\mathbf{C_2}|=1024$, there are $1,024^2=1,048,576$ unique combinations of the sub-codes.
Although the actual freedom of this \textit{conceptual} codebook is smaller than a real codebook with the same size, it already provides much larger capacity, given that we only maintain $|\mathbf{C_i}|\times k$ codes.
Prior arts reaches reconstruction saturation with codebook size $16,384$.
In Tab.~\ref{tab:tokenizer_ablation} of experiment, it is shown that factorizing the $32,768$ codebook into two $16,384$ sub-codebooks can further significantly improve the reconstruction performance.

\subsubsection{Disentanglement}
\label{sec:disentangle}
The factorized quantization design allows diverse feature learning, given the sufficient capacity in the feature adapters and sub-codebooks.
However, without explicit constraints, the sub-codebooks risk learning redundant and overlapping codes, particularly as the codebook size increases.
To address this issue, we propose a disentanglement regularization mechanism for the factorized sub-codebooks.

For simplicity, we take $k=2$ as an example scenario.
Through Eq.~\ref{eq:quant}, we obtain $q_1 \in \mathbb{R}^{L\times D}$ and $q_2 \in \mathbb{R}^{L\times D}$, where $L$ is the number of patches.
We design the disentanglement regularization mechanism as follows:
\begin{equation}
\mathcal{L}_{\text{disentangle}} = \frac{1}{n} \sum_{i=1}^{n} (q_1^\top q_2)^2,
\end{equation}
where $n$ is the number of samples in a batch.

This regularization mechanism minimizes the squared dot product between the two involved codes.
The dot product directly measures the affinity between the two codes after L2 normalization, ranging from $[-1, 1]$, where -1/1 indicates negative/positive correlation and 0 denotes orthogonality.
Minimizing the squaring function encourages the dot product value to approach 0.
It also provides a smooth gradient for optimization.
Note that this regularization does not directly apply to the entire codebook.
Instead, it operates on patches of each image instance.
In other words, for each patch, it encourages the involved sub-codes to capture different aspects.

\subsubsection{Representation Learning}
% With the disentanglement regularization, it is expected that the factorized codebooks would captures various different aspects of the image.
Typically, the main training objective of visual tokenizers is pixel reconstruction.
Research~\cite{lecun_recon} suggests that the reconstruction objective can hardly learn meaningful semantic features for perception, as the features mainly capture high-variance details.
However, recent work~\cite{saining_repa} finds that learning semantic features can benefit visual generation model training.
In this work, we show that representation learning plays a crucial role in tokenizer training, especially in the context of factorized quantization.

Consider the example of an image patch depicting an ear.
A traditional VQ code may capture its appearance, such as color, texture, etc.
However, it is unaware of the species, \textit{e.g.,} cat or dog.
While such a code may effectively reconstruct the patch, introducing semantic information is expected to be beneficial.
When informed with semantics, the decoder (and generation model) can better handle the corresponding visual reconstruction and generation tasks.
Moreover, compared to high-variance signals, semantic information tends to generalize better.

Building on this intuition, we introduce representation learning as a training objective to encourage the model to learn meaningful semantic features.
We continue to use $k=2$ as an example scenario.
Specifically, one sub-codebook, say $C_2$, is tasked with predicting the features of a pre-trained vision model using a lightweight feature prediction model.
$C_2$ essentially serves as the \textit{semantic} codebook that embeds the semantic information.
The other codebook $C_1$ functions as the \textit{visual} codebook that captures the visual details, complementing $C_2$.

We note that \textit{semantic} is still not a well-defined concept in the community.
As studied in the multimodal domain~\cite{tong2024cambrian}, pre-trained vision models place varying emphasis on the semantic property.
For instance, CLIP~\cite{radford2021learning}, which is pre-trained for cross-modal alignment, encodes high-level semantic features, while DINOv2~\cite{dinov2}, a self-supervised vision model, captures mid-level visual features.
Incorporating diverse vision models into the factorized sub-codebooks establishes a hierarchy of semantics: low-level structures (e.g., edges), mid-level details (e.g., textures), and high-level concepts (e.g., abstract appearance).

The total loss is a weighted sum of all the losses:
\begin{equation}
    \mathcal{L}_{\text{total}} = \mathcal{L}_{\text{VQGAN}} + \lambda_{1} \mathcal{L}_{\text{disentangle}} + \lambda_{2} \mathcal{L}_{\text{rep}},
\end{equation}
where $\lambda_{1}$ and $\lambda_{2}$ are weights.
In this paper, we present two variants of the implementation of FQGAN, including $k=2$ (\textbf{FQGAN-Dual}) and $k=3$ (\textbf{FQGAN-Triple}).
FQGAN-Dual employs CLIP~\cite{clip} as the pre-trained vision model to provide semantic features for the representation learning objective.
For FQGAN-Triple, CLIP~\cite{clip} and DINOv2~\cite{dinov2} are jointly adopted to form a semantic hierarchy.

\subsection{Auto-Regressive Model}
\label{sec:ar_model}
The factorized quantization design produces multiple sub-tokens for each spatial position, represented as $Z_t = (z_t^1, z_t^2, \dots, z_t^k)$, where $t$ denotes the time step.
Standard AR transformers, such as those in VQGAN~\cite{vqgan} and LlamaGen~\cite{llamagen}, predict only the index of the next token based on the hidden feature $g_t$, which makes them inherently unsuitable for handling factorized sub-tokens.
One simple solution is to apply $k$ classifiers to the hidden feature $g_t$, yielding the indices for the sub-tokens as $z_t^i = \text{cls}^i(g_t), i\in\{1,\dots,k\}$.
However, this method is shown to be suboptimal (see Tab.~\ref{tab:ar_head_ablation}).
To address this, we introduce a factorized AR head that sequentially predicts the distributions of these factorized sub-tokens, allowing for better modeling of their dependencies.
Fig.~\ref{fig:method} illustrates the full Factorized Auto-Regressive model (FAR). 
For each patch, the hidden feature $g_t$ serves as a prefix condition, which is processed by an additional AR head to autoregressively predict the list of sub-tokens, formulated as $z_t^i = \text{head}_{AR}(g_t; z_t^1, z_t^2, \dots, z_t^{i-1})$.
Following a scaling pattern similar to previous works~\cite{llamagen,openmatvit2}, FAR has Base and Larger versions, differentiated by their parameter sizes.
The detailed configurations are provided in the Appendix.

\section{Experiment}

\subsection{Setup}
In experiments, we follow previous works to use the standard benchmark, ImageNet~\cite{imagenet}, to train and evalaute the tokenizers and AR generation models.
For the factorization configuration, we experiment with $k=2$ and $k=3$.
$\lambda_{1}$ and $\lambda_{2}$ are empirically set to $0.1$ and $0.5$ respectively.
The training schedule of the visual tokenizer is adapted from LlamaGen~\cite{llamagen}.
Specifically, the tokenizer is trained with a global batch size of 256 and a constant learning rate of 2e-4 across 8 A100 GPUs.
For the AR model, we adopt a Llama-style~\cite{llama2,llamagen} transformer architecture as the backbone.
To accommodate the factorized codes, the model employs $k$ embedding layers on the input side, each embeds a separate sub-code, followed by a linear layer that aggregates these embeddings into a single representation.
On the output side, we adapt a factorized AR head that predicts the factorized codes for each patch.
The AR models are trained for 300 epochs with a constant learning rate of 2e-4 and a global batch size of 256 across 8 A100 GPUs.

\textbf{Metric.}
We adopt Fréchet inception distance (FID)~\cite{fid} as the main metric to evaluate visual tokenizers and generation models.
For tokenizers, we use the ImageNet validation set, consisting of 50k samples, to compute the reconstruction FID (rFID).
Additionally, we use PSNR and Inception Score~\cite{is} as auxiliary metrics for comparison.
For generation models, we follow the widely adapted ADM~\cite{adm} evaluation protocol to compute the generation FID (gFID).
Besides, Inception Score, Precision, and Recall are also used for comparison, following prior works.
In both quantitative and qualitative evaluations, we use classifier-free guidance~\cite{cfg} (CFG), with the weight set to 2.0.
We do not use any top-k or top-p sampling strategy unless specified.

\subsection{Comparison on Tokenizers}

\begin{table}[t]
    
    \centering
    \setlength{\tabcolsep}{4pt}
    \renewcommand\arraystretch{1.1}
    \caption{\textbf{Comparisons with other image tokenziers.} Reconstruction performance of different tokenizers on $256 \times 256$ ImageNet 50k validation set. All models are trained on ImageNet, except ``$*$" on OpenImages and ``$\dagger$" on unknown training data.
    \textbf{Bold} denotes the best scores; \underline{underline} denotes the second place.
    }
    \vspace{-5pt}
    \resizebox{\linewidth}{!}{
        \begin{tabular}{l|ccc|cc}
        \toprule
        \multirow{2}{*}{\textbf{Method}} & \textbf{Downsample} & \textbf{Codebook} & \textbf{Code} & \multirow{2}{*}{\textbf{rFID}$\downarrow$} & \multirow{2}{*}{\textbf{PSNR}$\uparrow$} \\
        
         & \textbf{Ratio} & \textbf{Size} & \textbf{Dim} & &  \\
        \midrule
        VQGAN~\citep{vqgan}   & 16 & 16384 & 256 & 4.98 & $-$  \\
        SD-VQGAN~\citep{ldm}  & 16 & 16384 & 4 & 5.15 & $-$  \\
        RQ-VAE~\citep{rqvae}  & 16 & 16384 & 256 & 3.20  &  $-$  \\
        LlamaGen~\citep{llamagen} & 16 & 16384 & 8 & 2.19 & 20.79 \\
        Titok-B~\citep{titok}  &  $-$ & 4096 & 12 &1.70 & $-$  \\
        VQGAN-LC~\citep{zhu2024scaling_vq} & 16 & 100000 & 8 & 2.62 & \textbf{23.80} \\
        VQ-KD~\citep{image_understand_tok} & 16 & 8192 & 32 & 3.41 & - \\
        VILA-U~\citep{vila_u} & 16 & 16384 & 256 & 1.80 & - \\
        Open-MAGVIT2~\cite{openmatvit2}  & 16 & 262144 & 1 & 1.17 & 21.90  \\
        \textbf{FQGAN-Dual}           & 16  & 16384 $\times$ 2 & 8 & 0.94 & 22.02  \\
        \textbf{FQGAN-Triple}   & 16  & 16384 $\times$ 3 & 8 &\textbf{0.76} & \underline{22.73}  \\ 
        \midrule
        SD-VAE$^\dagger$~\cite{ldm}    & 8  & 
        & 4 & 0.74 & 25.68  \\
        SDXL-VAE$^\dagger$~\cite{sdxl}  & 8  & $-$ & 4 & 0.68 & 26.04  \\ 
        \midrule
        ViT-VQGAN~\citep{vit-vqgan}   & 8 & 8192 & 32 & 1.28 & $-$  \\
        VQGAN$^{*}$~\citep{vqgan}   & 8 & 16384 & 4 & 1.19 & 23.38 \\
        SD-VQGAN$^{*}$~\citep{ldm}        & 8 & 16384 & 4 & 1.14 & $-$  \\
        OmniTokenizer~\citep{omnitokenizer}   & 8 & 8192 & 8 & 1.11 & $-$  \\
        LlamaGen~\citep{llamagen}   & 8 & 16384 & 8 & 0.59 & 25.45  \\
        Open-MAGVIT2~\citep{openmatvit2}   & 8 & 262144 & 1 & 0.34 & 26.19  \\
        \textbf{FQGAN-Dual}  & 8  & 16384 $\times$ 2 & 8 & \underline{0.32} & \underline{26.27}  \\
        \textbf{FQGAN-Triple}  & 8  & 16384 $\times$ 3 & 8 & \textbf{0.24} & \textbf{27.58}  \\ 
    \bottomrule
    \end{tabular}}
    \vspace{-15pt}
    \label{tab:main_tok_recon}
\end{table}

We first compare our method with popular visual tokenizers listed in Tab.~\ref{tab:main_tok_recon}.
Our FQGAN model sets a new state-of-the-art performance in discretized image reconstruction across various settings, including different codebook sizes and downsample ratios.
Compared to VQGAN and its advanced variants, our method outperforms them by a large margin.
Note our method is also built based on the vector-quantization mechanism.
This comparison effectively validates the advantage of our factorized quantization design.

Interestingly, compared to the state-of-the-art tokenizer Open-MAGVITv2, which employs an advanced lookup-free quantization mechanism, our method still exhibits superior image reconstruction performance, with a 0.41 rFID gap.
This result suggests that VQ-based methods still hold great potential for visual tokenization, which may have been overlooked previously.
Existing work often regards the codebook as a bottleneck, while our approach provides a novel perspective.
An explicit codebook offers the opportunity for more sophisticated designs on code embeddings, such as disentanglement and representation learning.

Another key finding is the comparison between SD-VAE~\cite{ldm}, SDXL-VAE~\cite{sdxl}, and our FQGAN.
SD-VAE and SDXL-VAE are advanced \textit{continuous} visual tokenizers widely used in Stable Diffusion models~\cite{sdxl,dit,dalle}.
We observe that our FQGAN, with a $16\times$ downsample ratio, achieves performance comparable to these continuous models, which use an $8\times$ downsample ratio.
In a fairer comparison, with both methods using an $8\times$ downsample ratio, our method achieves a significantly lower reconstruction FID, suggesting that \textit{discrete} representation in image tokenization is no longer a bottleneck for image reconstruction.

\subsection{Comparison on Generation Models}
\label{sec:compare_ar}
\begin{table}[t]
    \centering
    \setlength{\tabcolsep}{4pt}
    \renewcommand\arraystretch{1.0}
    \caption{
    \textbf{Class-conditional generation on $256 \times 256$ ImageNet.}
    Models with the suffix ``-re" use rejection sampling.
    The evaluation protocol and implementation follow ADM~\cite{adm}.
    Our model employs a cfg-scale of 2.0.
    }
    \resizebox{\linewidth}{!}{
    \begin{tabular}{c|l|c|cccc}
    \toprule
    \textbf{Type} & \textbf{Model} & \textbf{\#Para.} & \textbf{FID}$\downarrow$ & \textbf{IS}$\uparrow$ & \textbf{Precision}$\uparrow$ & \textbf{Recall}$\uparrow$  \\
    \midrule
    \multirow{4}{*}{\textcolor{gray}{Diffusion}} & \textcolor{gray}{ADM}~\citep{adm}  & \textcolor{gray}{554M}       & \textcolor{gray}{10.94} & \textcolor{gray}{101.0}        & \textcolor{gray}{0.69} & \textcolor{gray}{0.63}    \\
     & \textcolor{gray}{CDM}~\citep{cdm}   & \textcolor{gray}{$-$}       & \textcolor{gray}{4.88}  & \textcolor{gray}{158.7}       & \textcolor{gray}{$-$}  & \textcolor{gray}{$-$}   \\
     & \textcolor{gray}{LDM-4}~\citep{ldm} & \textcolor{gray}{400M}     & \textcolor{gray}{3.60}  & \textcolor{gray}{247.7}       & \textcolor{gray}{$-$}  & \textcolor{gray}{$-$}  \\
     & \textcolor{gray}{DiT-XL/2}~\citep{dit}  & \textcolor{gray}{675M}  & \textcolor{gray}{2.27}  & \textcolor{gray}{278.2}       & \textcolor{gray}{0.83} & \textcolor{gray}{0.57}   \\
    \midrule

    \multirow{2}{*}{\textcolor{gray}{LFQ AR}} & \textcolor{gray}{Open-MAGVIT2-B~\cite{openmatvit2}} & \textcolor{gray}{343M} & \textcolor{gray}{3.08} & \textcolor{gray}{258.26} & \textcolor{gray}{0.85} & \textcolor{gray}{0.51} \\
     & \textcolor{gray}{Open-MAGVIT2-L~\cite{openmatvit2}} & \textcolor{gray}{804M} & \textcolor{gray}{2.51} & \textcolor{gray}{271.70} & \textcolor{gray}{0.84} & \textcolor{gray}{0.54} \\
     \midrule
    
    \multirow{13}{*}{VQ AR} & VQGAN~\citep{vqgan} & 227M & 18.65 & 80.4         & 0.78 & 0.26    \\
     & VQGAN~\citep{vqgan}    & 1.4B   & 15.78 & 74.3   & $-$  & $-$     \\
     & VQGAN-re~\citep{vqgan}  & 1.4B  & 5.20  & 280.3  & $-$  & $-$     \\
     
     & ViT-VQGAN~\citep{vit-vqgan} & 1.7B & 4.17  & 175.1  & $-$  & $-$        \\
     & ViT-VQGAN-re~\citep{vit-vqgan}& 1.7B  & 3.04  & 227.4  & $-$  & $-$     \\
     
     & RQTran.~\citep{rqvae}       & 3.8B  & 7.55  & 134.0  & $-$  & $-$     \\
     & RQTran.-re~\citep{rqvae}    & 3.8B & 3.80  & 323.7  & $-$  & $-$    \\
     
     & LlamaGen-L~\citep{llamagen} & 343M & 3.80 & 248.28 & 0.83 & 0.51 \\
     & LlamaGen-XL~\citep{llamagen} & 775M & 3.39 & 227.08 & 0.81 & 0.54 \\
     
     & \textbf{FAR-Base}        & 415M  & 3.38 & 248.26 & 0.81 & 0.54 \\
     & \textbf{FAR-Large}       & 898M  & 3.08 & 272.52 & 0.82 & 0.54 \\
    \bottomrule
    \end{tabular}}
    \vspace{-2pt}
    \label{tab:main_gen}
\end{table}
We compare our FAR model with mainstream image generation models, including diffusion models, LFQ-based AR models, and VQ-based AR models, as shown in Tab.~\ref{tab:main_gen}.
Among VQ-based AR models, we observe that FAR achieves competitive image generation performance.
When comparing models with similar parameter sizes, specifically FAR-Base vs. LlamaGen-L and FAR-Large vs. LlamaGen-XL, our FAR model consistently achieves superior performance in both FID and Inception Score.
This validates the effectiveness of the proposed method.
Among the other methods, RQ-Transformer~\cite{rqvae} is similar to our method, as it also adopts an additional AR head to accommodate multiple sub-codes at each step.
The performance gap between RQ-Transformer and FAR further validates the power of our FQGAN tokenizer and its transferability to the downstream generation model.

When comparing FAR with Open-MAGVIT2~\cite{openmatvit2}, which shares a similar AR model design, our method exhibits a comparable or higher Inception Score, though with a slightly worse FID score.
The Inception Score suggests that our FQGAN tokenizer has the potential to match LFQ performance, while the FID score gap still demonstrates the superiority of LFQ compared to VQ, as studied in MAGVITv2~\cite{magvit2}.
Mitigating the \textit{generation} performance gap between LFQ and VQ is a critical yet challenging problem, which is beyond the scope of this work.
FQGAN is a crucial step toward this direction as it significantly improves image \textit{reconstruction} performance, surpassing both VQ and LFQ tokenizers.
Tab.~\ref{tab:main_gen} also suggests that the improvement on tokenization and reconstruction can be effectively transferred to AR generation.
We hope the FQGAN tokenizer will inspire related further research.

\begin{figure}[t!]
    \centering
    \includegraphics[width=0.95\linewidth]{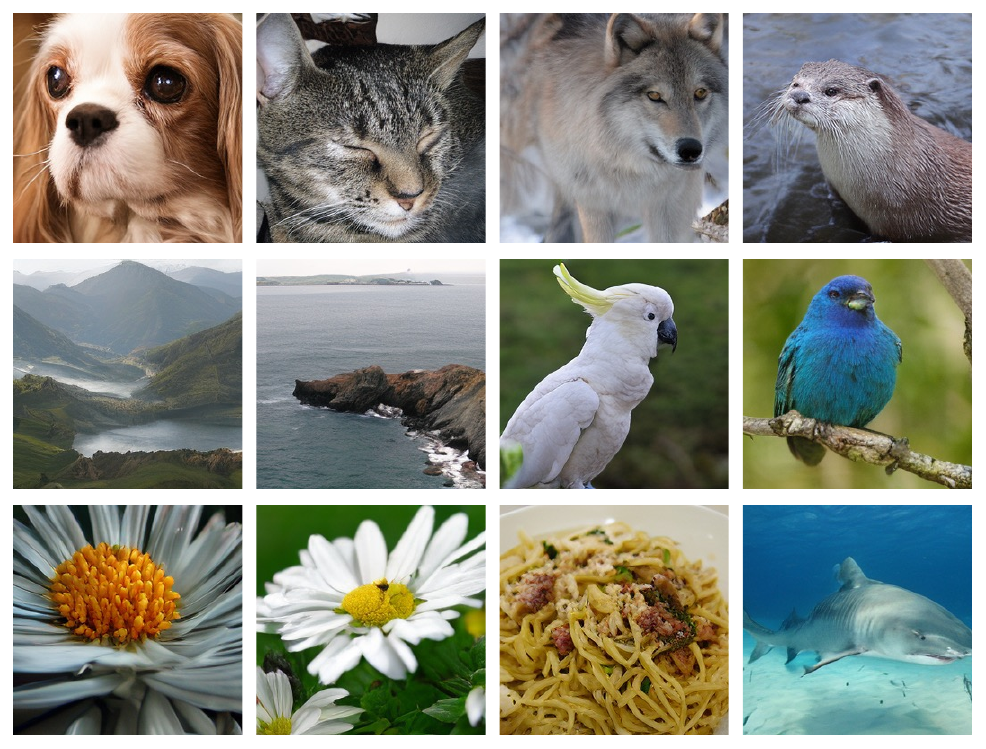}
    \vspace{-3pt}
    \caption{
    Qualitative examples generated by our FAR model.
    }
    \vspace{-4pt}
    \label{fig:ar_gen}
\end{figure}

Qualitative results of the FAR model is shown in Fig.~\ref{fig:ar_gen}.
The FAR model in this section is trained with tokens from FQGAN-Dual tokenizer.
More training details and settings of the FAR model are provided in the Appendix.

\subsection{Ablation Studies}

\begin{table}[t]
    \centering
    \setlength{\tabcolsep}{4pt}
    \renewcommand\arraystretch{1.2}
    \caption{
    Ablation study on different components of the proposed factorized quantization, using the FQGAN-Dual variant.
    }
    \vspace{-5pt}
    \resizebox{\linewidth}{!}{
        \begin{tabular}{l|ccc|cccc}
        \toprule
        \multirow{2}{*}{\textbf{Model}} & \textbf{Codebook} & \textbf{Dis.} & \textbf{Rep.} & \multirow{2}{*}{\textbf{rFID}$\downarrow$} & \multirow{2}{*}{\textbf{IS}$\uparrow$} & \multirow{2}{*}{\textbf{PSNR}$\uparrow$} & \multirow{2}{*}{\textbf{Usage}$\uparrow$} \\
         &  \textbf{Size} & \textbf{Regular.} & \textbf{Learn.} & & \\
        \midrule
        \multirow{2}{*}{VQGAN}  & 16384             & $-$ & $-$        & 3.71 & 50.05 & 20.56 & 98\% \\
                                & 32768             & $-$ & $-$        & 3.60 & 50.60 & 20.56 & 84\% \\
        \midrule
        \multirow{4}{*}{FQGAN}  & 16384 $\times$ 2  & \xmark & \xmark  & 2.00 & 54.72 & 22.21 & 97\% \\
                                & 16384 $\times$ 2  & \cmark & \xmark  & 1.84 & 55.04 & 22.04 & 98\% \\ 
                                & 16384 $\times$ 2  & \xmark & \cmark  & 1.73 & 55.00 & 21.61 & 98\% \\ 
                                & 16384 $\times$ 2  & \cmark & \cmark  & 1.66 & 55.21 & 21.62 & 98\% \\ 
    \bottomrule
    \end{tabular}}
    \vspace{-5pt}
    \label{tab:tokenizer_ablation}
\end{table}

\paragraph{Factorized Quantization.}
We investigate the design components of the FQGAN tokenizer, including the factorized codebook, disentanglement regularization mechanism, and representation learning objective.
In this study, we adopt FQGAN-Dual, \textit{i.e.,} $k=2$.
All experiments are conducted for 10 epochs on the ImageNet training set to ensure a fair comparison.
As shown in Tab.~\ref{tab:tokenizer_ablation}, we start with a vanilla VQGAN tokenizer.
Increasing the codebook size from $16,384$ to $32,768$ results in a drop in codebook usage, yielding only marginal performance gains even with double codebook size.
Previous studies~\cite{llamagen} have shown that with training schedule of more epochs, the $32,768$ version ultimately performs worse than the $16,384$ version.
Next, we consider a vanilla factorized codebook design, which splits the single $32,768$ codebook into two $16,384$ sub-codebooks.
Such factorization brings a significant performance gain, as reflected in the rFID score change from $3.60$ to $2.00$.
Compared to a single codebook with the same number of codes (\textit{i.e.}, capacity), the factorized design greatly reduces lookup complexity.
It also yields more diverse code combinations, improving performance by a large margin.

\paragraph{Disentanglement and Representation Learning.}
Next, we gradually incorporate the proposed additional designs into the factorized codebooks.
By making only one change in each experiment, we find that both the disentanglement regularization and the representation learning objective lead to better reconstruction results.
When applied together, the two designs achieve even better performance.
We attribute this performance gain to the fact that disentanglement regularization forces the factorized codes to learn more diverse aspects of the image, while the representation learning objective explicitly incorporates semantic information.
It is worth mentioning that an $\text{rFID}=2.0$ for the vanilla factorization version is already a very strong result, rarely achieved by previous VQ tokenizers.
Pushing the performance further is particularly challenging, which effectively demonstrates the strength of the proposed designs.

\paragraph{What has each sub-codebook learned?}

\begin{figure}[t!]
    \centering
    \includegraphics[width=0.98\linewidth]{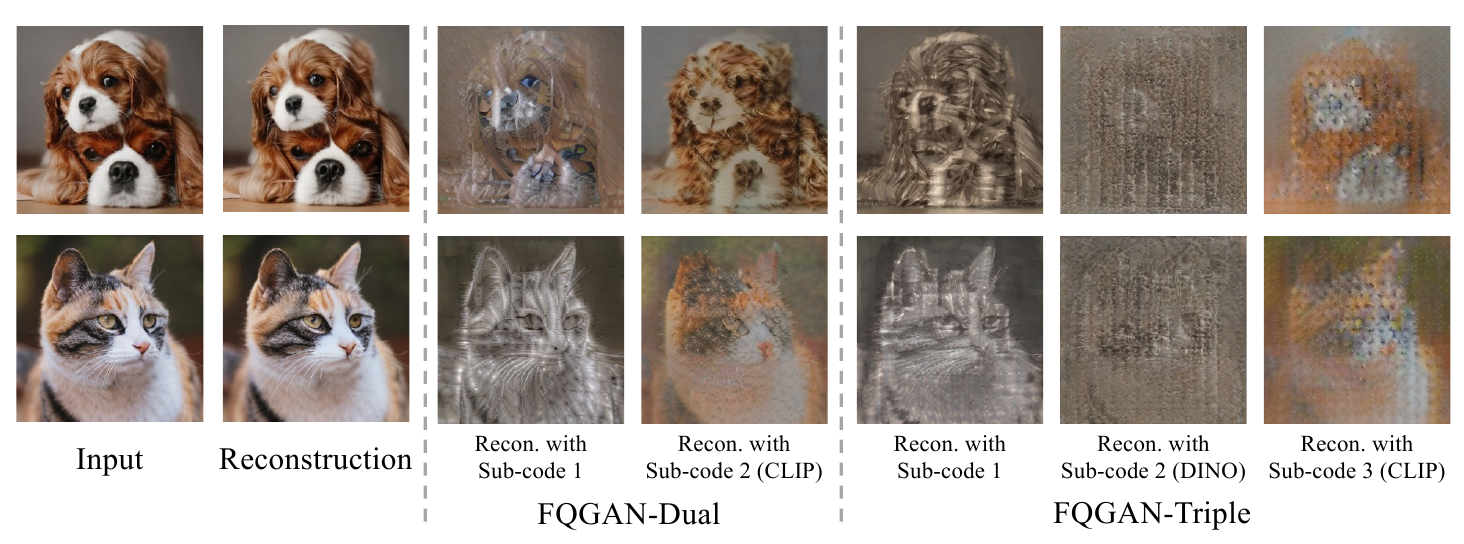}
    \vspace{-8pt}
    \caption{
    Visualization of standard reconstruction by FQGAN-Dual and reconstruction using only a single sub-codebook.
    }
    \label{fig:vis_recon_codebook}
    \vspace{-5pt}
\end{figure}

To better understand the underlying behavior of the factorized sub-codebooks, we provide a comprehensive visualization.
Fig.~\ref{fig:vis_recon_codebook} demonstrates the reconstruction results, including standard reconstruction and reconstruction using only a single sub-codebook, achieved by setting the rest of the code embeddings to zero.
In the two sub-codebooks of FQGAN-Dual, we observe that sub-codebook 1 highlights low-level features, such as essential structures, shapes, and edges of the image.
Sub-codebook 2, jointly supervised by CLIP features, presents a high-level abstract version of the original image, where colors are blurred together, and textures are preserved in a softened manner.
When factorizing further into three sub-codebooks, \textit{i.e.,} FQGAN-Triple, we observe that sub-codebook 1 still emphasizes the low-level strong edges and overall shape.
Sub-codebook 2, jointly supervised by DINO features, highlights textural elements, preserving surface patterns and fine details without clear structural outlines, representing mid-level features.
Finally, sub-codebook 3 concentrates on higher-level appearance and produces an abstract or blurry version of the original image.
This visualization suggests that the factorized sub-codebooks are indeed tasked with capturing different aspects of the image.
With the supervision of representation learning, the sub-codebooks naturally form complementary hierarchical levels of visual features.

\begin{figure}[t]
    \centering
    \includegraphics[width=0.9\linewidth]{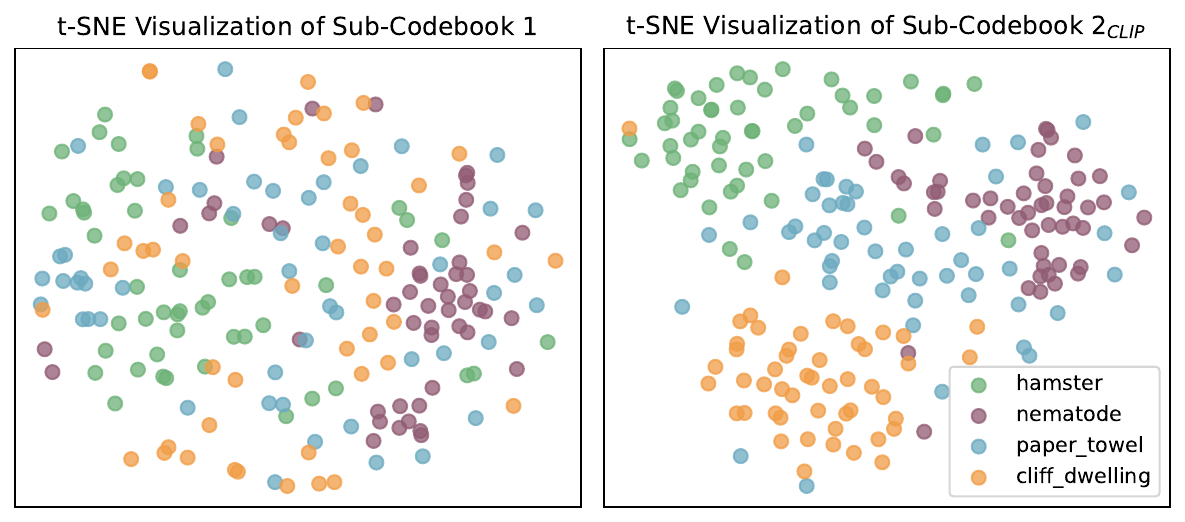}
    \vspace{-8pt}
    \caption{
    T-SNE visualization of VQ codes from different sub-codebooks in FQGAN-Dual.
    }
    \label{fig:vis_tsne_dual}
    \vspace{-10pt}
\end{figure}

Furthermore, we illustrate the distribution of VQ codes from different sub-codebooks.
Following previous practice~\cite{image_understand_tok}, we randomly sample four classes from the ImageNet dataset, encode them with our tokenizer, and visualize the distribution using the t-SNE technique.
The left part of Fig.~\ref{fig:vis_tsne_dual} shows that VQ codes from sub-codebook 1, without additional regularization, are distributed in an unordered manner in the space.
This is likely because this sub-codebook is solely trained for reconstruction, capturing high-variance detail while lacking awareness of semantic categories.
In contrast, the right part suggests that the CLIP-supervised sub-codebook 2 exhibits better semantic awareness, as its codes from the same category are distributed within a cluster.
The two visualizations effectively demonstrate what each sub-codebook has learned qualitatively.
We provide more visualizations in the Appendix.

\paragraph{Effect of AR Head.}
\begin{table}[t]
    \centering
    \setlength{\tabcolsep}{4pt}
    \renewcommand\arraystretch{1.}
    \caption{
    Ablation study on the generation model head design with the proposed FQGAN tokenizer. We use FAR-Large model with cfg-scale=1.75 in this study.
    }
    \vspace{-5pt}
    \resizebox{0.93\linewidth}{!}
    {
        \begin{tabular}{l|c|c}
        \toprule
        \textbf{Generation Model Head}  & \textbf{Top-k Sampling}  & \textbf{gFID}$\downarrow$ \\
        \midrule
        \multirow{2}{*}{$k$ Linear Classifiers}  & 4096 & 5.19   \\
                                                 & 8192 & 6.90   \\  
        \midrule
        \multirow{2}{*}{$k$ MLP Classifiers}     & 4096 & 5.59   \\
                                                 & 8192 & 8.88   \\ 
        \midrule
        \multirow{2}{*}{Factorized AR Head}      & 4096 & 4.37   \\ 
                                                 & 8192 & 3.74   \\ 
    \bottomrule
    \end{tabular}}
    \vspace{-4pt}
    \label{tab:ar_head_ablation}
\end{table}

Adapting the FQGAN tokenizer to auto-regressive visual generation models presents the challenge of handling multiple sub-codes at each step.
This is crucial, as predicting a wrong sub-code at a specific position can invalidate the entire patch.
We present this investigation in Tab~\ref{tab:ar_head_ablation}.
We begin with a simple solution that employs $k$ independent linear classifier heads to decode the hidden embedding of the AR backbone into their respective sub-codebooks in parallel.
This strategy yields decent results but lags behind auto-regressive models with the same parameter level.
We hypothesize that this is due to the parallel decoding scheme placing too heavy a burden on the classifier.
Therefore, we attempt to increase the capacity of the classifier by using multiple layers with a non-linear activation function in between.
However, as shown in the table, the MLP version performs even worse, suggesting that simply increasing the capacity and computation is not the key to addressing this issue.

In factorized auto-regressive generation, the key issue is that the mismatch between sub-codes within a position (patch) can significantly affect the results.
This suggests that an effective design is a module that not only decodes from the AR backbone but also models the dependency between sub-codes.
To this end, we explore using an additional auto-regressive head to decode the factorized sub-codes.
The last row of Tab.~\ref{tab:ar_head_ablation} shows that this design can improve performance by a considerable margin.
For example, when decoding code $z_t^2$, the vanilla classifier or MLP version only references the hidden embedding $g_t$ output by the AR backbone, whereas the AR module allows the decoding process to also attend to code $z_t^1$, strengthening the dependency among sub-codes of the current patch and improving overall generation quality.

\section{Discussion and Future Work}
In this work, we design a factorized quantization method and explore dual and triple sub-codebooks.
Future research on factorizing more sub-codebooks could be a promising direction.
Secondly, since the sub-codebook is jointly supervised by strong vision models, such as CLIP, it is interesting to probe its performance on multimodal understanding tasks.
We provide a preliminary exploration in the Appendix.
In the long term, building a truly unified tokenizer that excels at both generation and understanding tasks would be beneficial to the community.
We believe the factorized design is a promising direction toward this ultimate goal, as it entails various levels of visual information.
Regarding limitations, as discussed in Sec.~\ref{sec:compare_ar}, our method outperforms previous VQ-based methods in both reconstruction and generation.
However, in downstream generation, our model still lags behind LFQ-based methods in generation FID metric.
Our work, with a strong reconstruction performance, serves as an initial step toward bridging the gap between VQ and LFQ.
We hope this work inspires future research to push the boundary further.

\section{Conclusion}
We focus on a critical limitation of current VQ tokenizers: their difficulty in handling large codebooks.
To address this, we propose a novel factorized quantization approach that decomposes a large codebook into multiple independent sub-codebooks.
To facilitate learning of the sub-codebooks, we design a disentanglement regularization mechanism that reduces redundancy while promoting diversity.
Additionally, we introduce a representation learning objective that explicitly guides the model to learn meaningful semantic features.
The proposed visual tokenizer, FQGAN, effectively handles large codebooks and achieves state-of-the-art performance in discrete image reconstruction, surpassing both VQ and LFQ methods.
Experimental results show that this tokenizer can be integrated into auto-regressive image generation models by adding a factorized AR head, demonstrating competitive image generation performance.
Besides, we provide an in-depth analysis to unveil how the factorized codebooks function.
Finally, we discuss several limitations to inspire future works.

\section*{Appendix}
\appendix

\section{More Generation Results}

\begin{figure*}[p]
    \centering
    \includegraphics[width=0.9\linewidth]{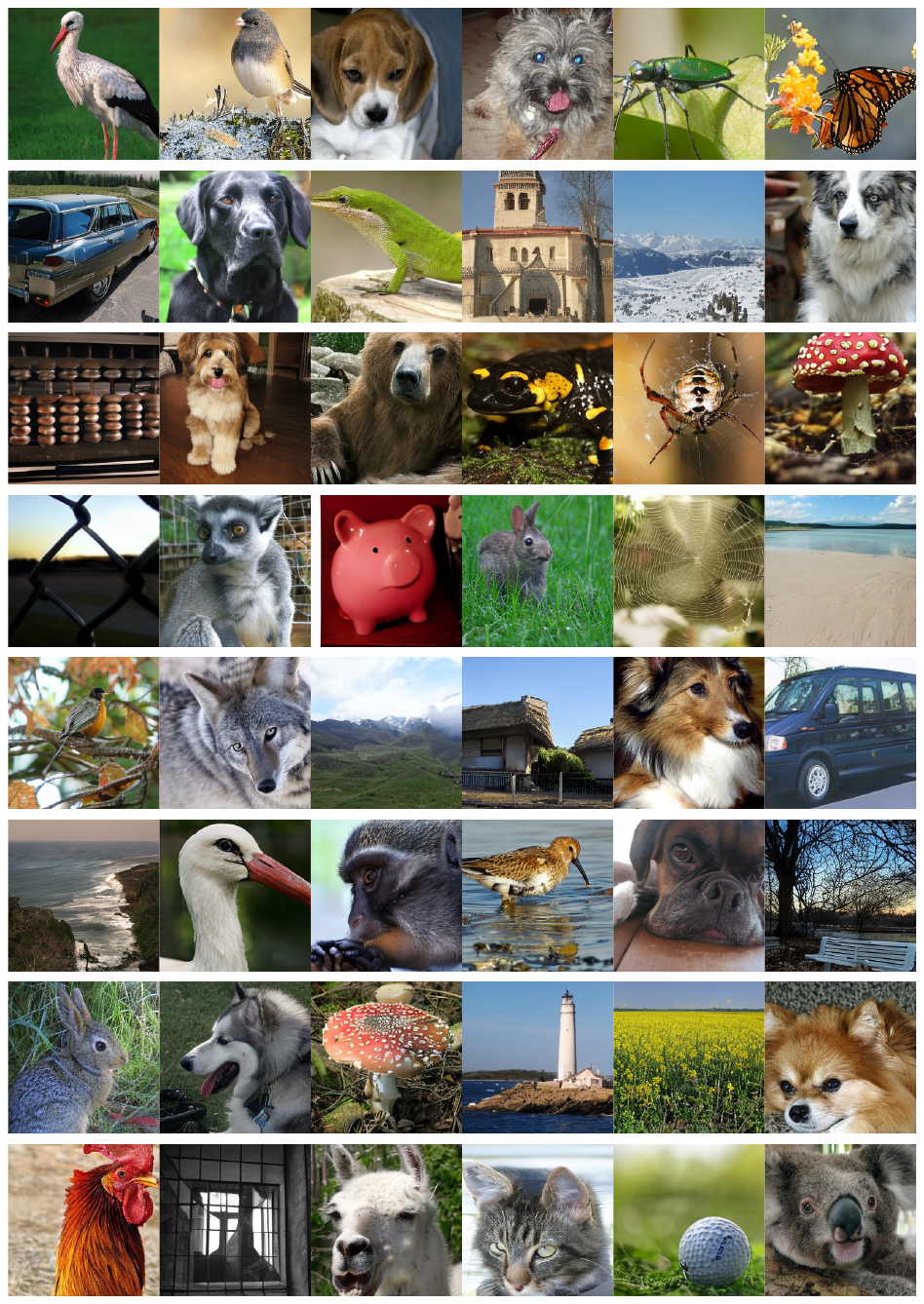}
    \caption{
    More qualitative examples generated by the FAR model.
    }
    \label{fig:ar_gen_supp}
\end{figure*}

Figure~\ref{fig:ar_gen_supp} presents additional examples generated by our FAR model, highlighting its impressive image generation capabilities.

\section{More Training Details of Visual Tokenizers}

In this section, we demonstrate the flexibility of the proposed FQGAN in terms of extending its codebook size and scaling its training schedule.
Table~\ref{tab:tokenizer_ablation_epoch} provides detailed experimental results for both FQGAN-Dual ($k=2$) and FQGAN-Triple ($k=3$).
We observe that increasing the number of sub-codebooks from $2$ to $3$—effectively raising the total codebook size from $16384\times 2$ to $16384\times 3$—further improves reconstruction quality.
With only 10 epochs of training, FQGAN-Triple achieves an rFID of 1.30, outperforming the FQGAN-Dual variant under the same training conditions.
We attribute the performance gain to the larger codebook ($16384\times 3$), which introduces additional capacity, and the factorization design and associated training objectives enrich the new sub-codebook with more diverse features.

We observe that training the tokenizer for only 10 epochs does not fully utilize the large capacity of the sub-codebooks.
To address this, we extend the training schedule to further explore the capacity of the model.
As shown in Tab.~\ref{tab:tokenizer_ablation_epoch}, increasing the training epochs from 10 to 40 significantly enhances performance.
FQGAN-Dual improves from an rFID of 1.66 to 0.94, while FQGAN-Triple achieves an rFID of 0.76, comparable to the performance of \textit{continuous} features.
This study suggests that the FQGAN model has significant potential for scaling to achieve improved performance, owing to its factorization design.
Importantly, training for 40 epochs does not indicate saturation.
Due to limited time and resources, we did not extend training beyond 40 epochs; however, additional training could potentially yield even lower rFID values.

\section{Extended Analysis of Sub-codebooks}

\begin{table}[t] \small
    \centering
    \setlength{\tabcolsep}{1.5mm}
    \caption{
    The proposed FQGAN is extendable to multiple codebooks, i.e., $k>2$, and demonstrate scaling behavior with increasing training schedule.
    }
    \resizebox{0.95\linewidth}{!}
    {
        \begin{tabular}{c|cc|ccc}
        \toprule

        $\mathbf{k}$ & \textbf{Codebook Size} & \textbf{Epoch} & \textbf{rFID}$\downarrow$ & \textbf{IS}$\uparrow$ & \textbf{PSNR}$\uparrow$ \\
         
        \midrule
        \multirow{3}{*}{2}   & 16384 $\times$ 2  & 10   & 1.66 & 55.21 & 21.62 \\
                             & 16384 $\times$ 2  & 20   & 1.25 & 56.39 & 22.00 \\  
                             & 16384 $\times$ 2  & 40   & 0.94 & 57.15 & 22.02 \\
        \midrule
        \multirow{3}{*}{3}  & 16384 $\times$ 3   & 10   & 1.30 & 56.41 & 21.85 \\ 
                            & 16384 $\times$ 3   & 20   & 0.92 & 57.80 & 22.67 \\ 
                            & 16384 $\times$ 3   & 40   & 0.76 & 58.05 & 22.73 \\ 
    \bottomrule
    \end{tabular}}
    \label{tab:tokenizer_ablation_epoch}
\end{table}

\begin{figure}[t!]
    \centering
    \includegraphics[width=0.95\linewidth]{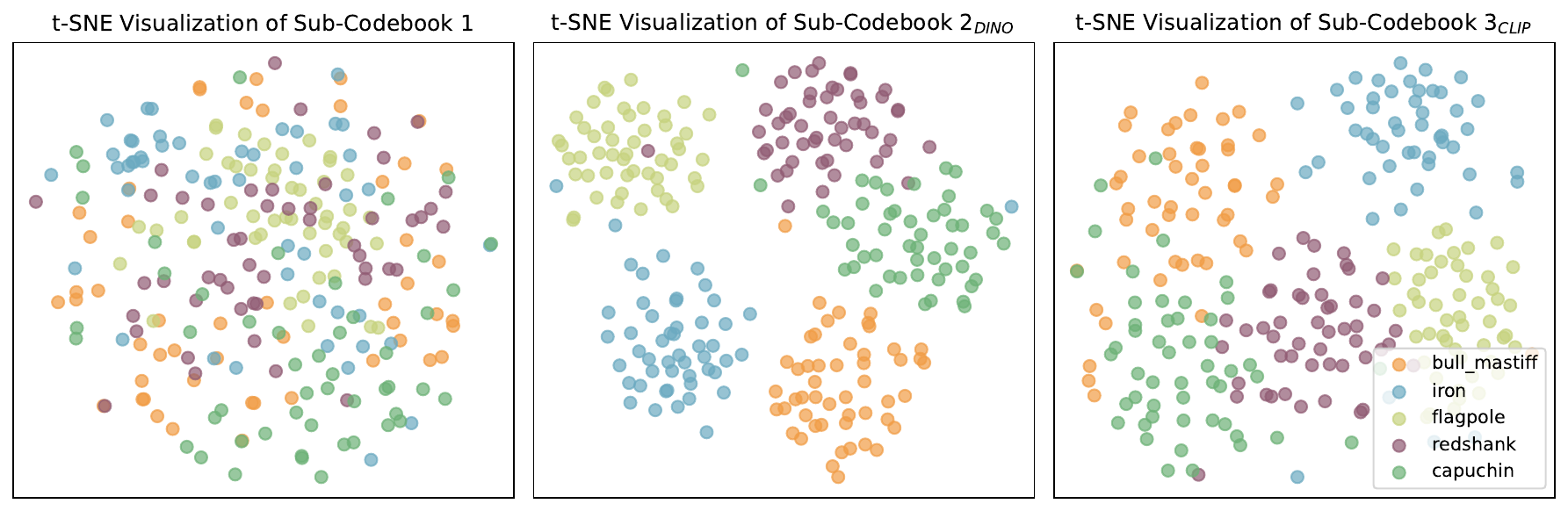}
    \caption{
    T-SNE visualization with FQGAN-Triple.
    }
    \label{fig:tsne_triple}
\end{figure}

We present a visualization of the distribution of VQ code embeddings for the FQGAN-Triple model in Fig.~\ref{fig:tsne_triple}.
Specifically, FQGAN-Triple is equipped with three factorized sub-codebooks.
Sub-codebook 2 is jointly supervised using DINOv2 features, while sub-codebook 3 is jointly supervised using CLIP features.
Following prior practice~\cite{image_understand_tok}, we sample five classes from the ImageNet dataset and use the FQGAN-Triple model to encode these images into the latent space.
Then, we use the t-SNE technique to visualize the code embeddings from each sub-codebook.
We observe that the code embeddings from sub-codebook 1 appear unordered in the space, likely due to the dominant influence of high-variance details.
This observation is consistent with the visualization of FQGAN-Dual in the main paper.
In contrast to sub-codebook 1, the other two sub-codebooks are organized into clusters based on image categories.
This clustered distribution likely reflects the influence of the representation learning objective.

Interestingly, we observe that sub-codebook 2 is distributed in a more compact manner compared to sub-codebook 3, with embeddings from the same category clustering closer to the center.
To better understand this phenomenon, we investigate the specific model instances and their performance on ImageNet.
Specifically, we use \texttt{facebook/dinov2-small} and \texttt{openai/clip-vit-base-patch16} checkpoints as the vision models.
The DINOv2 checkpoint achieves $81.1\%$ linear probing accuracy on the ImageNet validation set, while the CLIP checkpoint achieves $68.3\%$ accuracy.
This performance gap likely explains the observed differences in the visualization.
The CLIP model is designed to capture cross-modal information between vision and language, while DINOv2 performs better in vision-centric classification tasks.
These differing objectives lead the FQGAN-Triple model to naturally form a semantic hierarchy.

\clearpage
\clearpage

\section{Exploration on Multimodal Understanding}

With the representation learning objective, the factorized sub-codebooks learn a semantic hierarchy, spanning from visual details to high-level concepts.
Recent studies on unified multimodal models~\cite{lwm,showo,vila_u} demonstrate that their multimodal understanding performance is largely limited by traditional visual tokenizers.
Compared to a standard tokenizer trained with a reconstruction objective, FQGAN demonstrates greater potential for supporting multimodal understanding tasks.
We conduct a preliminary experiment to investigate its potential.

Specifically, we use a LLaVA~\cite{llava}-style architecture with a Phi~\cite{abdin2024phi} LLM as the base model.
The traditional LLaVA model undergoes two-stage training.
In stage-1, a projector is trained to connect a vision model with the LLM for cross-modal alignment.
In stage-2, the projector and LLM are trained jointly to develop instruction-following capabilities.
In this study, we train only the stage-1 projector to evaluate the potential of vision features for cross-modal alignment.
Subsequently, we evaluate the model on the Flickr-30k~\cite{plummer2015flickr30k} test set using a simple image captioning task.
The results are shown in Tab.~\ref{tab:llava_mmu}.

Firstly, when comparing continuous features extracted from the VQ encoder to discrete code embeddings from the codebook, we observe that continuous features consistently perform better.
This suggests that continuous features are still more suitable for cross-modal understanding, as they contain richer information.
Secondly, when comparing different sub-codebooks, sub-codebook 2 consistently outperforms sub-codebook 1 in the captioning task.
This demonstrates that joint supervision with CLIP features enhances the cross-modal potential of sub-codebook 2.
We further observe that combining the two sub-codebooks results in comparable or better performance, particularly in the captioning task with continuous features.
This phenomenon suggests that the visual details from sub-codebook 1 have the potential to complement information missing in sub-codebook 2, enabling more effective cross-modal alignment and understanding.

\begin{table}[]
    \centering
    \setlength{\tabcolsep}{4pt}
    \renewcommand\arraystretch{1.0}
    \caption{
    Exploration on multimodal understanding with LLaVA~\cite{llava} framework and Flicker-30k~\cite{plummer2015flickr30k} benchmark.
    }
    \resizebox{0.65\linewidth}{!}{
    \begin{tabular}{cc|c}
    \toprule
    \textbf{Feature}  & \textbf{Sub-Codebook} & \textbf{CIDEr} \\
    \midrule
    \multirow{3}{*}{Continuous} & Codebook 1 & 3.67 \\
                                & Codebook 2 (CLIP) & 7.15 \\
                                & Both & 10.28  \\
    \midrule
    \multirow{3}{*}{Discrete}   & Codebook 1 & 2.22 \\
                                & Codebook 2 (CLIP) &  7.40 \\
                                & Both & 7.37  \\
    \bottomrule
    \end{tabular}}
    \label{tab:llava_mmu}
\end{table}

The performance metrics of our model remain significantly lower than those of standard captioning models, likely due to the following factors.
First, in our FQGAN model, the feature dimension is compressed to 8, which is significantly smaller than the typical dimensions of traditional vision features (512 or 768).
While t-SNE visualizations demonstrate clear category separability, the features possess less semantic richness compared to the continuous representations generated by a standard vision backbone.
Increasing the feature dimension further could enhance performance by capturing more detailed semantic information.
Second, using CLIP features as auxiliary supervision introduces only a limited amount of cross-modal semantic information into the VQ encoder, especially given the relatively small scale of the training dataset.
The VILA-U~\cite{vila_u} study suggests that initializing the model with pre-trained CLIP weights could be a promising approach.
However, it also highlights that training a single codebook to simultaneously optimize for reconstruction and semantic objectives can lead to feature conflicts.

Given these considerations, our FQGAN model shows strong potential to advance multimodal understanding and contribute to the development of unified multimodal frameworks.
The factorized sub-codebook design effectively mitigates feature conflicts between high-variance visual details and high-level semantic concepts, naturally establishing a hierarchical structure.
We hope this study serves as a foundation for further research in this area.

\begin{figure*}[h]
    \centering
    \begin{subfigure}{0.4\textwidth}
        \centering
        \includegraphics[width=\linewidth]{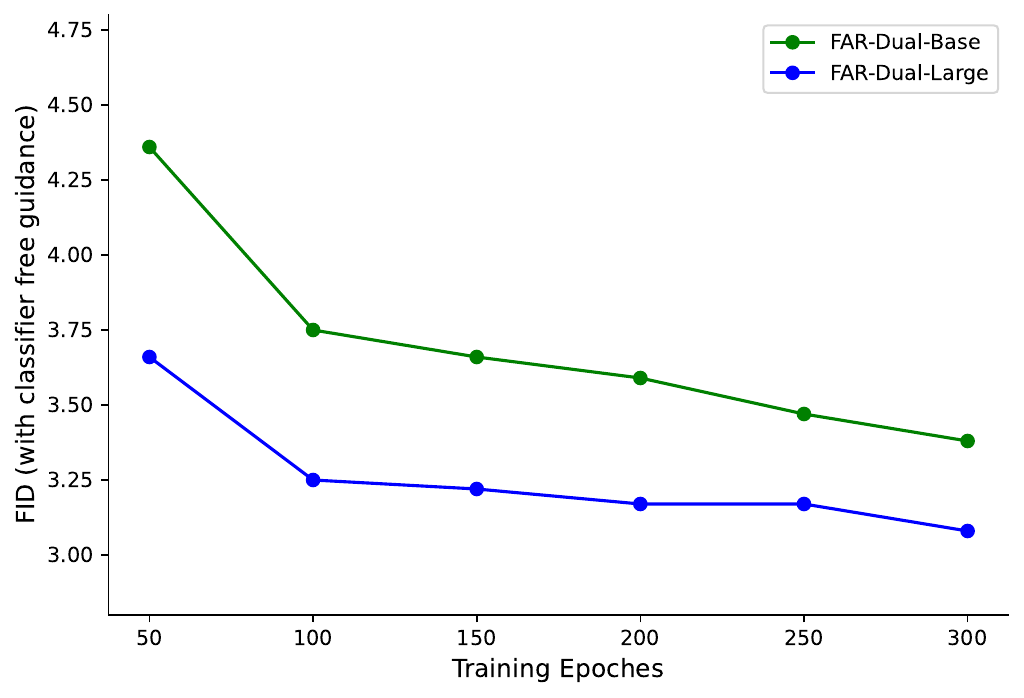}
        \caption{FAR-Dual generation FID with CFG.}
    \end{subfigure}
    \hfill
    \begin{subfigure}{0.4\textwidth}
        \centering
        \includegraphics[width=\linewidth]{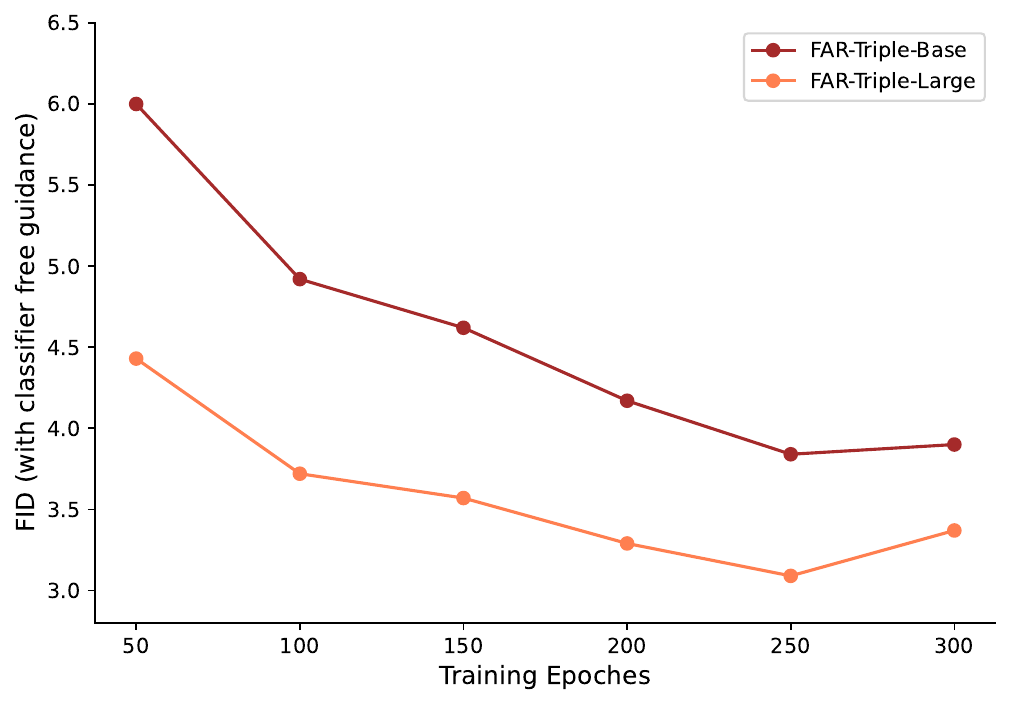}
        \caption{FAR-Triple generation FID with CFG.}
    \end{subfigure}
    \caption{Training details of the FAR model. We demonstrate FAR-Dual and FAR-Triple with both Base and Large size.}
    \label{fig:scaling_curve}
\end{figure*}

\section{More Training Details of AR Generation Models}
\begin{table}[]
    \centering
    \setlength{\tabcolsep}{4pt}
    \renewcommand\arraystretch{1.0}
    \caption{\textbf{Model configurations of FAR}. We partially follow the scaling rule proposed in the previous work~\citep{llamagen}.}
    \resizebox{\linewidth}{!}{
    \begin{tabular}{cccccc}
    \toprule
    \textbf{Model}  & \textbf{Parameters} & \textbf{AR Backbone} & \textbf{AR Head} &  
    \textbf{Widths}  & \textbf{Heads} \\
    \midrule
    FAR-Base & 415M & 24 & 3 & 1024 & 16 \\
    FAR-Large & 898M & 36 & 4 & 1280 & 20 \\
    \bottomrule
    \end{tabular}}
    \label{tab:scaling_model_size}
\end{table}

Table~\ref{tab:scaling_model_size} presents the detailed configurations of the FAR-Base and FAR-Large models.
The AR backbone and AR head architecture follows a standard auto-regressive transformer design with causal attention.

Next, we present the training details of the auto-regressive generation model in Fig.~\ref{fig:scaling_curve}.
Specifically, we plot the gFID score curves using classifier-free guidance (cfg=2.0 for FAR-Dual and cfg=1.75 for FAR-Triple).
Firstly, we observe that FAR scales well across different model sizes, with larger models consistently achieving better FID scores, regardless of whether dual or triple codes are used.
Next, when comparing FAR-Dual and FAR-Triple models with the same number of parameters, we observe that FAR-Dual achieves a lower gFID score than FAR-Triple.
For the ``-Large" model size, FAR-Dual and FAR-Triple achieve comparable best gFID scores: $3.08$ vs. $3.09$.
For the ``-Base" model size, FAR-Dual outperforms FAR-Triple, achieving gFID scores of $3.38$ vs. $3.84$.
This performance gap suggests that handling multiple sub-codes in auto-regressive generation models remains challenging.
The ablation study on the AR head in the main paper suggests that further scaling the AR head size could improve learning performance.
We leave this for future work due to limited computational resources.

\section{Discussion on Related Works}
Our work is closely related to some existing studies.
Residual Quantization~\cite{rqvae} and Modulated Quantization~\cite{zheng2022movq} implicitly adopt the philosophy of factorization by decomposing visual features into primary and residual or modulated components.
While this factorization improves image reconstruction performance, its potential is limited by reliance on a single codebook.
Our approach explicitly decomposes a large codebook into multiple independent sub-codebooks, introducing greater flexibility and efficiency.

A concurrent work, ImageFolder~\cite{imagefolder}, introduces a visual tokenization approach using two codebooks to encode semantics and details, achieving improvements in reconstruction quality. However, our work differs significantly with ImageFolder in objectives, representation design, generative modeling strategy, and downstream applicability. Below, we summarize the key differences:

\begin{itemize}
\item  Objective and Focus.  
   ImageFolder focuses on improving the computational efficiency of autoregressive image generation through a Folded Tokenization mechanism, which compresses spatial information to reduce sequence length. Its primary goal is to optimize high-resolution image generation by addressing scalability challenges.  
   In contrast, our work emphasizes building interpretable and disentangled visual representations through Factorized Tokenization, prioritizing semantic clarity and downstream usability. Our framework is designed not only for efficient generation but also for achieving more meaningful and structured representations.

\item Representation Design.
   While ImageFolder employs two codebooks to separately encode semantics and details, it does not enforce explicit independence between these codebooks. Instead, its tokenization primarily focuses on spatial compression to maintain efficiency.  
   In our work, we explicitly enforce independence between codebooks, disentangling semantic and detail representations. Furthermore, we introduce a hierarchical multi-codebook structure, enabling richer and more interpretable visual representations that support a broader range of tasks and downstream applications.

\item Generative Modeling Strategy.
ImageFolder uses an autoregressive model to directly generate folded tokens, focusing on maintaining spatial coherence during generation. Its decoding relies on sequential predictions of compressed tokens.
In contrast, our work explores how to effectively transfer the factorized tokenizer into downstream autoregressive generation tasks. By leveraging a factorized autoregressive prediction head, our framework enhances the generation process, enabling high-quality and consistent outputs. This demonstrates the adaptability of our approach for autoregressive generation tasks and highlights its advantages in maintaining both semantic integrity and visual fidelity.

\item  Evaluation and Applicability.  
   The evaluation in ImageFolder primarily measures reconstruction quality and generation efficiency using metrics like FID and inference latency.
   Our work takes a broader perspective, assessing the interpretability and usability of factorized representations. While our primary focus is on enhancing autoregressive generation, we also evaluate how the disentangled representations enable structured representation learning and multimodal understanding. This highlights the versatility of our approach in scenarios requiring more nuanced and interpretable models.

\item Role of Pre-trained Vision Models.  
   Both works utilize pre-trained vision models for supervision or regularization, but the integration differs. ImageFolder leverages these models to improve feature extraction for reconstruction tasks.  
   In our framework, pre-trained models are utilized to guide the disentanglement process, enabling the creation of a hierarchical and factorized tokenization structure. This approach enhances the adaptability and generalizability of our representations to diverse downstream tasks.

\item Summary.
While both ImageFolder and our work aim to improve visual tokenization and generation, their focus and methodologies diverge significantly. ImageFolder prioritizes computational efficiency and scalability in tokenized image generation, whereas our work introduces explicit disentanglement and hierarchical factorization for autoregressive generation. These innovations establish a more interpretable and versatile framework, extending the potential applications of visual tokenization beyond reconstruction and sequence efficiency to tasks requiring semantically meaningful and structured representations.

\end{itemize}

Together, these works, including our FQGAN, highlight factorization as a promising avenue for advancing visual tokenization and generation.

{
    \small
    \bibliographystyle{ieeenat_fullname}
    \bibliography{main}
}

\end{document}